\documentclass[conference]{IEEEtran}
\IEEEoverridecommandlockouts
\usepackage{cite}
\usepackage{url}
\usepackage{amsmath,amssymb,amsfonts}
\usepackage{algorithmic}
\usepackage{graphicx}
\usepackage{textcomp}
\usepackage{xcolor}
\usepackage{times}
\usepackage{latexsym}
\usepackage{graphicx}
\usepackage{multirow}
\usepackage{color}
\usepackage{array}
\usepackage{float}
\usepackage{booktabs}
\usepackage{tabularx}
\usepackage{amsmath}
\usepackage{amsfonts,amssymb}
\usepackage{CJKutf8}
\def\BibTeX{{\rm B\kern-.05em{\sc i\kern-.025em b}\kern-.08em
    T\kern-.1667em\lower.7ex\hbox{E}\kern-.125emX}}
\begin{document}

\title{CNMBERT: A Model for Converting Hanyu Pinyin Abbreviations to Chinese Characters}

% \author{\IEEEauthorblockN{Zishuo Feng$^{1}$, Feng Cao$^{2}$, Lei Feng$^{1, \dagger}$}
\author{\IEEEauthorblockN{Zishuo Feng$^{1}$, Feng Cao$^{2}$}
\IEEEauthorblockA{\textit{$^{1}$ Beijing University of Posts and Telecommunications, Beijing China}}
\IEEEauthorblockA{\textit{$^{2}$ Beihang University, Beijing China}}
\IEEEauthorblockA{\textit{AkatukiFZS@bupt.edu.cn, Kohaku@buaa.edu.cn}}
% \IEEEauthorblockA{\textit{$^{\dagger}$ Corresponding author: fenglei@bupt.edu.cn}}
}
% \author{\IEEEauthorblockN{Anonymous Authors}}
\maketitle

\begin{abstract}
The task of converting Hanyu Pinyin abbreviations to Chinese characters is a significant branch within the domain of Chinese Spelling Correction (CSC). It plays an important role in many downstream applications such as named entity recognition and sentiment analysis. This task typically involves text-length alignment and seems easy to solve; however, due to the limited information content in pinyin abbreviations, achieving accurate conversion is challenging. In this paper, we treat this as a fill-mask task and propose CNMBERT, which stands for zh-\textbf{CN} Pinyin \textbf{M}ulti-mask \textbf{BERT} Model, as a solution to this issue. By introducing a multi-mask strategy and Mixture of Experts (MoE) layers, CNMBERT outperforms fine-tuned GPT models and ChatGPT-4o with a 61.53\% MRR score and 51.86\% accuracy on a 10,373-sample test dataset.
\end{abstract}


\section{Introduction}
\par
\begin{CJK*}{UTF8}{gbsn}
Nowadays, on social media platforms using Simplified Chinese like Douyin\footnote{\url{https://en.wikipedia.org/wiki/TikTok##Douyin}}, people often replace certain Chinese words with the first letter of their pinyin. For instance, people may use “b l” which stands for “暴力(violent)”. This is especially common when users wish to avoid prohibited words or reduce the effort of typing long pinyin sequences. One of the most notable manifestations of this phenomenon is the “B-culture” on Douyin\footnote{\url{https://www.zhihu.com/question/269016377/answer/2654824753}}. To maintain a positive and healthy atmosphere on the platform, Douyin is hypercorrecting by implementing numerous ridiculous prohibited words. Many ordinary words have been flagged as prohibited. As shown in Fig. 1, users are not allowed to use the word “病 (disease)” in videos or comments as it may lead to deletion by the platform's automated system. Therefore, they came up with using “b” (the pinyin's first letter of “病”) as a substitute for it.
\par
To better understand this phenomenon, it is important to explain pinyin. Pinyin, also known as Hanyu Pinyin, is the romanization system for Simplified Chinese\footnote{\url{https://en.wikipedia.org/wiki/Pinyin}} . It represents the pronunciation of a Chinese character, like the phonetic symbols in English(e.g., \textit{“bao li” for “暴力(violent)”}).  
\end{CJK*}
\par
\begin{figure}
    \centering
    \includegraphics[width=1\linewidth]{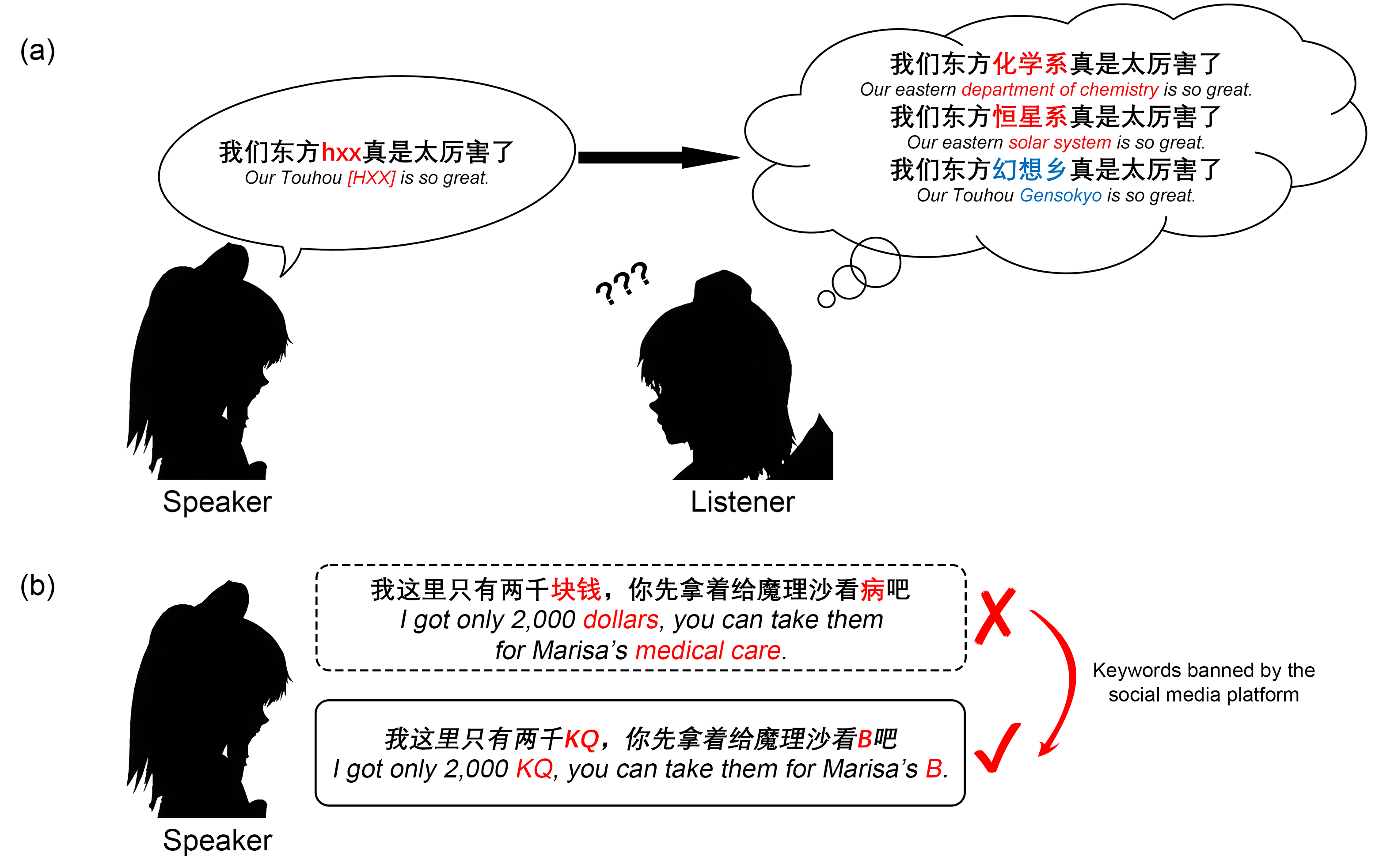}
    \begin{CJK*}{UTF8}{gbsn}
    \caption{(a). A simple example of pinyin Abbreviation can make listener confused. The correct characters are highlighted in blue. In this context, “hxx” refers to “幻想乡(Gensokyo)”. (b). The example of some platforms may designate words related to money and illness as prohibited words. }
    \end{CJK*}
    \label{fig:enter-label}
\end{figure}
This phenomenon, where the pinyin's first letters are used to replace Chinese characters, is referred to as \textbf{pinyin abbreviations}, an important branch of Chinese Spelling Correction (CSC) \cite{b1}. With the advent of the transformer architecture \cite{b2}
, an increasing number of GPT models \cite{b3} have been released. These models have shown remarkable proficiency in understanding the semantics of Chinese text and have been widely applied to tasks such as machine translation, text generation, and sentiment analysis. However, these models face challenges when dealing with pinyin abbreviations. First, they lack sufficient training on pinyin abbreviation data. Second, pinyin abbreviations convey minimal information, making them hard to predict. Moreover, the same abbreviation can correspond to multiple pinyin forms, each potentially linked to numerous homophones. As shown in Table I, this ambiguity complicates the accurate interpretation of abbreviations, even for native Chinese speakers. The difficulty arises from both the limited informational content of abbreviations and the complex nature of Chinese phonetics.
\begin{CJK*}{UTF8}{gbsn}
\begin{table}
    \centering
    \renewcommand{\arraystretch}{1.2}
    \caption{A simple example of pinyin abbreviation to character conversion task on different models and their predictions ordered by probability. The correct characters are highlighted in red and the incorrect characters are highlighted in blue.}
    \resizebox{\linewidth}{!}{
    \begin{tabular}{cc}
    \hline
    \multirow{2}*{Source} & 因为学业繁重，所以我\textcolor{red}{\textbf{fq}}了音乐 \\
    ~ & \textit{Due to a heavy academic workload, I \textcolor{red}{[fq]} music.} \\
    \multirow{2}*{Target} & 因为学业繁重，所以我\textcolor{blue}{\textit{\textbf{放弃}}}了音乐 \\ 
    ~ & \textit{Due to a heavy academic workload, I \textcolor{blue}{gave up on} music.} \\ 
    \hline
        \multirow{5}*{ChatGPT-4o} & 因为学业繁重，所以我\textcolor{red}{\textbf{发圈}}了音乐\\
        ~ & \textit{Due to the heavy academic workload, I'm \textcolor{red}{\textbf{circling}} the music} \\
        \cline{2-2}
        ~ & 因为学业繁重，所以我\textcolor{red}{\textbf{发起}}了音乐 \\
        ~ & \textit{Due to the heavy academic workload, I \textcolor{red}{\textbf{initiated}} music} \\ 
       \hline
        \multirow{5}*{Qwen2.5-72b} & 因为学业繁重，所以我\textcolor{red}{\textbf{翻墙}}了音乐 \\
        ~ & \textit{Due to the heavy academic workload, I'm \textcolor{red}{\textbf{using VPN}} music} \\
        \cline{2-2}
        ~ & 因为学业繁重，所以我\textcolor{red}{\textbf{放起}}了音乐 \\ 
        ~ & \textit{Due to the heavy academic workload, I'm \textcolor{red}{\textbf{playing}} music} \\ 
       \hline
        \multirow{5}*{CNMBERT} & 因为学业繁重，所以我\textcolor{blue}{\textbf{放弃}}了音乐 \\
        ~ & \textit{Due to a heavy academic workload, I \textcolor{blue}{gave up on} music.} \\
        \cline{2-2}
        ~ & 因为学业繁重，所以我\textcolor{red}{\textbf{放起}}了音乐 \\ 
        ~ & \textit{Due to the heavy academic workload, I'm \textcolor{red}{\textbf{playing}} music} \\ 
    \end{tabular}
    }

    \label{tab:my_label}
\end{table}
\end{CJK*}
\par
In this paper, based on the Whole Word Mask Chinese BERT model (Chinese-BERT-wwm) \cite{b4}, we propose CNMBERT\footnote{\url{https://github.com/IgarashiAkatuki/CNMBert}}
, which stands for zh-\textbf{CN} Pinyin \textbf{M}ulti-mask \textbf{BERT} Model
to solve the pinyin abbreviation to character conversion task. We treat this as a fill-mask task and extend the number of mask tokens in the simple BERT model \cite{b5} to match the number of letters in the pinyin alphabet. Each letter is mapped to a distinct mask token (\textit{e.g., The letter 'a' will be mapped to mask token \textbf{[LETTER\_A]}} ), then the model predicts the corresponding mask tokens, generating the translation of pinyin abbreviation. Since BERT is pre-trained on the fill-mask task, this ensures consistency between pre-training and downstream tasks. We also replaced a portion of the model's Feed-Forward Networks (FFNs) with Mixture of Experts (MoE) layers, which allows the model to process the information in pinyin abbreviations better. In the subsequent experimental sections, we will demonstrate that both strategies are effective.
\par
We tested the model on a test dataset we constructed, which includes 10,373 pinyin abbreviation translation data points. The results indicate that CNMBERT better understands and utilizes the information conveyed by pinyin abbreviations. It significantly outperforms our baseline model trained with Qwen2.5-14b-Instruct \cite{b6} and surpasses the performance of GPT-4o, achieving state-of-the-art (SOTA) performance. It also provides potential performance improvement solutions for downstream tasks, such as named entity recognition \cite{b7} and sentiment analysis \cite{b8}.

\section{Related Work}
In this section, we will discuss related work associated with our model, as well as work on pinyin translation.
\subsection{BERT}
Currently, most methods regard CSC as a sequence tagging task and fine-tune BERT models \cite{b9}. Compared to GPT models, BERT uses bidirectional attention. Due to it being an autoencoder model and the use of the Masked Language Modeling (MLM) task as its pretraining task, BERT can fully understand the context, making it generally perform better than GPT models on grammatical error correction tasks \cite{b5}.
\par
The original BERT model uses the WordPiece tokenizer \cite{b10} to segment the input into WordPiece tokens. In this case, a single word could be split into several smaller pieces. When training on the fill-mask task, those words may only have part of their tokens masked. For Chinese, the original BERT model segments input by characters, with each character mapped to an individual token. 
\par
However, this approach can also result in only part of a polysyllabic word (a word composed of multiple Chinese characters) being masked. To address the issue of only parts of words being masked, the Google team proposed Whole Word Mask (WWM). This approach allows the model to mask entire words' tokens, making it easier for the model to predict the masked content \cite{b5}. For the Chinese BERT model, there is a version called Chinese-BERT-wwm \cite{b4}. The input text is first segmented into words using LTP, a Chinese text segmentation tool \cite{b11}. Subsequently, when BERT selects words to be masked, the entire word will be masked rather than just a single character within it if a part of a polysyllabic word is chosen. Experimental results have shown that the BERT model using WWM outperforms the original BERT model, achieving SOTA performance on most tasks.
\begin{CJK*}{UTF8}{gbsn}
\begin{table*}
    \centering
    \renewcommand{\arraystretch}{1.2}
    \caption{Examples of different Masking Strategies. Our multi-mask Strategy uses the first letter of pinyin to replace the default [MASK] token. The pinyin of words “一无是处” and “梦想” are “\textbf{Y}i \textbf{W}u \textbf{S}hi \textbf{C}hu” and “\textbf{M}eng \textbf{X}iang”.} 
    \resizebox{\linewidth}{!}{
    \begin{tabular}{cc}
    \hline
       \multirow{2}*{Masking Strategy} & 即使一无是处的人也可以谈梦想吗 \\
       ~ & \textit{(Can even worthless people talk about their dreams?)} \\
       \hline
       \multirow{2}*{Original Masking}  &  即 使 一 无 是 \textbf{[MASK]} 的 人 也 可 以 谈 梦 \textbf{[MASK]} 吗 ?\\
       ~ & \textit{(Can even \textbf{[MASK]} \#\#less people talk about their \textbf{[MASK]} \#\#eams ?)} \\
       \cline{2-2}
       \multirow{2}*{$+$WWM} & 即 使 \textbf{[MASK] [MASK] [MASK] [MASK]} 的 人 也 可 以 谈 \textbf{[MASK] [MASK]} 吗 ? \\
        ~ & \textit{(Can even \textbf{[MASK] [MASK]} people talk about their \textbf{[MASK] [MASK]} ?)} \\
       \cline{2-2}
        \multirow{2}*{$+$Multi-Mask (Our)}  &  即 使 \textbf{[Y] [W] [S] [C]} 的 人 也 可 以 谈 \textbf{[M] [X]} 吗 ? \\
        ~ & \textit{(Can even \textbf{[Y] [W] [S] [C]} people talk about their \textbf{[M] [X]} ?)} \\
    \end{tabular}
    }
    \label{tab:my_label}
\end{table*}
\end{CJK*}
\subsection{PinyinGPT}
\par
\begin{CJK*}{UTF8}{gbsn}
PinyinGPT is a model based on a publicly available character-level GPT-2 model \cite{b12}, specifically designed to address GPT-2's limited performance when handling pinyin sequences and abbreviations. While the GPT model can often infer the meaning of perfect pinyin (\textit{e.g., “dian nao” is the perfect pinyin for “电脑”}) sequences from context, its performance drops significantly when handling pinyin abbreviations \cite{b13}. 
\end{CJK*}
\par
To solve this problem, the authors propose PinyinGPT-Concat and PinyinGPT-Embed, adding pinyin information to the input sequence. Specifically, PinyinGPT-Concat appends the pinyin sequence of input Chinese characters to their context. During inference, the model's input format is \textbf{$[w_1, ..., w_n, [SEP], p_{n+1}, ..., p_{n+k}, [SEP]]$}, where [SEP] is a special token used to separate the text from the pinyin, with \textbf{$p_{n}$ }representing the pinyin of \textbf{$w_1$}. PinyinGPT-Embed, on the other hand, introduces an additional pinyin embedding layer, which performs word embeddings on the first letters of the pinyin.
\par
Experimental results showed that PinyinGPT outperforms the standard GPT model in handling pinyin abbreviation inputs. However, PinyinGPT is trained to serve as part of an input method, aiming to predict the meaning of pinyin sequences and abbreviations at the end of a sentence. Thus, from the design perspective of PinyinGPT, it is a text generation task that generates possible Chinese characters based on a given sequence or abbreviation of pinyin at the end of a sentence. The results also show that PinyinGPT can only translate abbreviations at the end of sentences and lacks the ability to translate abbreviations within sentences.
\section{Model}
In this section, we will provide a detailed description of the architecture of our CNMBERT and explain how we attempt to enable the BERT model to understand and leverage the information contained in pinyin abbreviations.

\subsection{Multi-Mask Strategy}
\par
The original BERT model is pre-trained on the Masked Language Modeling (MLM) and Next Sentence Prediction (NSP) tasks. In the MLM task, 15\% of the characters in the input sentence are replaced with the [MASK] token, and the model is required to predict the masked words based on the surrounding context. Pinyin abbreviations carry limited information, making it challenging for models to utilize them effectively. Although the MLM task trains the model to predict [MASK] tokens, the original [MASK] tokens lack any informational content. To address this limitation, we propose an enhanced MLM task by introducing the multi-mask strategy, where each [MASK] token is enriched with information related to the pinyin abbreviations. 
\par
Therefore, building upon the WWM approach, we modified the masking logic. Specifically, we expanded the number of mask tokens so that each character is mapped to a distinct mask token. During training, our model first selects the words to be masked. Then, using the pypinyin package\footnote{\url{https://github.com/mozillazg/python-pinyin}}, we retrieve the first letter of the pinyin for each character. This letter is mapped to the corresponding mask token, which serves as the masked token for that character, as illustrated in Table II. The multi-mask strategy can be applied to any BERT-based model, endowing it with the capability to convert pinyin abbreviations to Chinese characters.
\par
In this way, we “embedded” the first letter of pinyin into the mask tokens, allowing them to carry this information. During the prediction phase, this information can constrain the model's output, making it more likely to generate Chinese characters whose first letter matches those of the provided pinyin abbreviation, as shown in Table III. 

\par
To address the mismatch between the pre-training task and downstream tasks, the original BERT model introduces a mechanism where a small portion of selected words are randomly replaced with random characters or left unchanged, in order to enhance the model's robustness. Since CNMBERT pre-training task aligns with the downstream tasks, we have opted to discard this adjustment. In the model's pre-training phase, all selected words follow the multi-mask strategy. Furthermore, we increased the probability of masking polysyllabic words rather than individual characters. This strategy encourages the model to focus more on training with polysyllabic words, thereby enhancing its prediction capability for such words.
\begin{CJK*}{UTF8}{gbsn}
\begin{table}
    \centering
    \renewcommand{\arraystretch}{1.2}
     \caption{After using the multi-mask strategy, the CNMBERT can effectively predict relevant Chinese words based on the first letter of the pinyin. In this table, we have bolded the first letter of the pinyin.}
    \resizebox{\linewidth}{!}{
    \begin{tabular}{cc}
        \hline
       Input & 现在网上一堆无良\textcolor{blue}{\textbf{mt}}歪曲事实，吃人血馒头 \\
       \cline{2-2}
       \multirow{2}*{Target} & 现在网上一堆无良\textcolor{blue}{\textbf{媒体}}歪曲事实，吃人血馒头
       \\
        ~ & \textit{(There are so many unscrupulous \textcolor{blue}{\textbf{media}} online distorting the facts.)} \\
        \cline{2-2}
       Pinyin & 媒体 (\textbf{M}ei \textbf{T}i)\\
       \hline
        \multirow{3}*{Prediction} & 媒体 \textbf{M}ei \textbf{T}i (Media) \\
         ~ & 模特 \textbf{M}o \textbf{T}e (Model)  \\
         ~ & 木头 \textbf{M}u \textbf{T}ou (Wood) \\
    \end{tabular}
    }
    \label{tab:my_label}
\end{table}
\end{CJK*}
\par
Due to the change in the masking strategy, the loss function for the MLM task in the model also underwent a modification. We define the input sequence as $x=(x_1, x_2, x_3, ..., x_n)$, the model's output sequence as $y=(y_1, y_2, y_3, ..., y_n)$, and the set of masked indices as $M =\{M_a, M_b, M_c, ..., M_z\}$, where the $M_a$ is the index set of the characters masked as \textbf{\textit{[LETTER\_A]}}. For our CNMBERT model, the cross-entropy loss function is defined as:

\begin{align}
    \mathcal{L}_{MLM}= -\sum_{S\in M}\sum_{i \in S} logP(y_i|x_i)
\end{align}

\begin{figure*}
    \centering
    \includegraphics[width=0.85\linewidth]{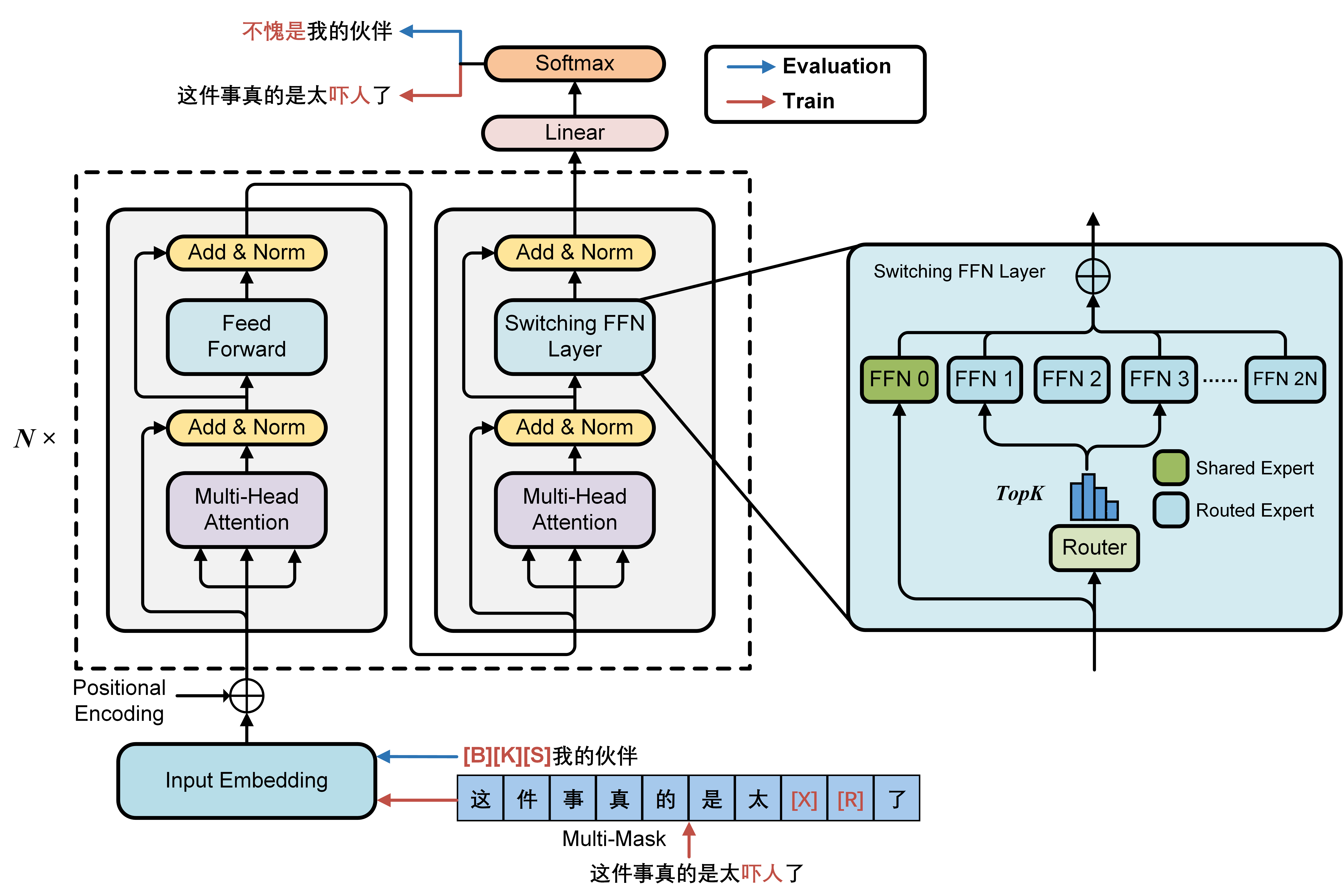}
    \caption{The overall architecture of the model and its workflow. The model use a 16-layer transformer architecture. For layers 0–15, some FFN layers are replaced with MoE layers, each containing a shared expert. Specifically:
In layer [1, 3, 5], there are \textbf{2} experts and \textbf{top-k = 1}.
In layers [7], there are \textbf{4} experts and \textbf{top-k = 1}.
In layers [9, 11, 13, 15], there are \textbf{8} experts and \textbf{top-k = 2}.
And for other layers using the regular FFN.}
    \label{fig:enter-label}
\end{figure*}
\subsection{MoE Layer}
\par
In CNMBERT, some FFN layers are replaced with MoE layers \cite{b14}. This approach allows the model to scale up and enables different experts to handle different tokens. The MoE layers in our model are designed as shown in Fig. 2, based on the Pyramid-Residual Mixture of Experts (PR-MoE) proposed by DeepSpeed \cite{b15}. It has fewer parameters compared to a sparse MoE but achieves similar performance. Additionally, it has fewer experts in the MoE layers close to the input, so the MoE layers in the model form a pyramid-like structure. For our model, each MoE layer includes one shared expert, with the rest as routed experts \cite{b16}. The MoE layers replace every other feedforward layer. Additionally, the number of routed experts activated during inference varies across different layers. In the shallower layers, we use fewer routed experts and activate a smaller number of experts during inference.

\par
In the model, the final output of the MoE layer is a linear combination of the outputs from the shared expert and the routed experts. For a token $x$, the output is defined as follows, where $L(x): \mathbb{R}^{dim(x)} \xrightarrow{} \mathbb{R}^{num\_experts}$, $TopK_{N}(W)$ selects the top N values in tensor $W$. $a$ and $b$ are the model's trainable parameters.
\begin{equation}
        \begin{split}
    Output(x) = \alpha \cdot SharedExpert(x) \\ 
    + \beta \cdot Routed(x)\
    \end{split}
\end{equation}
\begin{align}
    \alpha,\beta &= Softmax([a, b]) \\
     W &= Softmax(L(x)) \\
     Routed(x)  &= \sum_{w_i \in  TopK_{N}(W)}w_i \cdot Expert_i(x)
\end{align}

\par
\section{Experiments}
In this section, we will describe the training details for both the baseline model and CNMBERT, the dataset used to train and evaluate the model, as well as the results of the pinyin abbreviation to character conversion task.

\subsection{Datasets}
We use the EXT dataset
\footnote{\url{https://github.com/ymcui/Chinese-BERT-wwm}},
webtext2019zh dataset 
\footnote{\url{https://github.com/brightmart/nlp_chinese_corpus}}
\cite{b27} and the Bilibili (a Chinese video-sharing platform) comment dataset for training and evaluating the model. These datasets cover a wide range of topics, including mathematics, history, society, and various other domains. Additionally, these data are sourced from Chinese Q\&A websites, Chinese Wikipedia, and Chinese streaming media services, encompassing both standard written Chinese expressions and non-standard internet Chinese expressions. This ensures diverse data sources and reduces potential bias in the training data. Then we stratified and extracted 2 million data points from these datasets for training the model. 
\par
\begin{table}[H]
    \centering
    \renewcommand{\arraystretch}{1.5}
    \caption{Data distribution of test dataset.}
    \resizebox{\linewidth}{!}{
    \begin{tabular}{c|ccccc}
    \hline
    \multirow{2}*{Word Length} & \multirow{2}*{Total} & \multirow{2}*{Monosyllabic Words} & \multicolumn{3}{c}{Polysyllabic Words} \\
    \cline{4-6}
    ~ & ~ & ~ & 2 & 3 & 4+ \\
    \hline
       Rate & 100\% & 32.57\% & 59.86\% & 6.18\% & 1.39\% \\
       Count & 10,373 & 3,379 & 6,209 & 641 & 144 \\
       Different Replaced Words & 4,967 & 701 & 3,553 & 577 & 136 \\
    \end{tabular}
    }
    \label{tab:my_label}
\end{table}
To construct the test dataset shown in Table IV, we randomly selected 10,373 sentences that were different from the training data. These sentences were segmented using pkuseg \cite{b18}, and in each sentence, we randomly chose one word and replaced it with its pinyin's first letter. Additionally, we manually reviewed those data to ensure that the replaced words would not introduce excessive ambiguity. Then, to ensure the diversity of the test dataset, we also imposed constraints on the frequency of the different replaced words, ensuring that their replacement frequency in the overall dataset is less than 0.4\%.

\subsection{Baseline Model and Pre-traing}
In our work, we developed a baseline model based on 
Qwen2.5-14b-Instruct \cite{b6} . We fine-tuned the model using a specifically created dataset of pinyin abbreviations, which enables it to understand and process tasks related to the interpretation and translation of pinyin abbreviations more effectively.
We applied Low-Rank Adaptation (LoRA) \cite{b17} to fine-tune the foundational Qwen model. Specifically, we adjusted the query (\textbf{\textit{Q}}) and value (\textbf{\textit{V}}) weight parameter matrices in each transformer layer to better adapt the model to the task of pinyin abbreviation to character. For the baseline model, we fine-tuned it for 3 epochs using LlamaFactory \cite{b19} and Deepspeed ZeRO-3 \cite{b20}, with an initial learning rate of 5e-5. To distinguish it from the original model, we refer to the fine-tuned model as Qwen-FT in the following text.
\par
We extended the BERT-wwm-ext model \cite{b4} by increasing its layers from 12 to 16, and the parameters of these additional layers are randomly initialized. Then we fine-tuned the model on the original MLM task using the aforementioned data with the WWM strategy. Based on this, we trained three CNMBERT models: (a) a model without multi-mask strategy and MoE, (b) a model with multi-mask strategy, and (c) a model with both multi-mask strategy and MoE. For the MoE model, the weights of the shared experts were copied from the FFN layers of our fine-tuned BERT-wwm-ext. Both models were trained using the same data and with a learning rate of 1e-5 with a linear warmup scheduler. We ran the warmup process for 1 epoch and trained 32 epochs in total.
% \par
%  Specifically, for each transformer layer, we fine-tuned the query and value matrices by LoRA, while freezing the other parameters. 
\par
All models were trained using the Transformers package \cite{b21} with the AdamW optimizer \cite{b22}, and the training was conducted on a server with four Nvidia RTX 4090 GPUs.

% \subsection{Evaluation}

\subsection{Evaluation Metric}
As mentioned earlier, a pinyin abbreviation can correspond to multiple pinyin forms, and each pinyin form may represent multiple homophones. As a result, pinyin abbreviations can sometimes be inherently ambiguous. In certain specific contexts, the multiple words corresponding to a pinyin abbreviation are all contextually appropriate. Therefore, we use mean reciprocal rank (MRR) as the evaluation metric and additionally conduct separate tests for monosyllabic and polysyllabic words. The calculation of MRR at top-K is defined as follows: it measures whether the ground truth exists within the top-K results and identifies its position.

\begin{equation}
    MRR@K = \frac{1}{N}\sum_{i=1}^{N}\mathbb{I}(rank_i \leq K)\cdot \frac{1}{rank_i}
\end{equation}

\subsection{Additional Settings}
We evaluated our CNMBERT, ChatGPT-4o, Qwen2.5-14B-Instruct \cite{b6}, Qwen-FT, and Llama3-Chinese-8B-Instruct \cite{b24} on a test environment with 1 Nvidia RTX 3090 Ti GPU. Both LLaMA3 and Qwen2.5 were deployed using Ollama\footnote{\url{https://ollama.com/}} and accessed via LangChain \cite{b25}. 
\par
For all CNMBERT models, we use beam search to generate prediction results with hyperparameters $beam\_size = 16$ and $topk = 10$. At each time step, we apply softmax to the probabilities of the current results.
\par
For all GPT models, we use the same prompt, which is as follows:
\begin{CJK*}{UTF8}{gbsn}
~\\
\hrule
~\\
\par
\textcolor{red}{\textbf{System:}} \textit{Do your best to answer the user's question to the best of your ability. Avoid repeating the question in your response. Do not use redundant or repetitive statements. Ensure your language is smooth and coherent. Avoid saying something once and then repeating it later unnecessarily. Do not use emojis. Please answer in Chinese.} 
\par
\textcolor{blue}{\textbf{User:}} \textit{In the sentence [他们就这样xf地居住在一起], [xf] represents the first letter of the pinyin for certain Chinese characters. What could it possibly mean? Please list all the potential meanings in descending order of likelihood, separated by spaces. Do not provide any additional explanatory text.}
\par
\textcolor{blue}{\textbf{Model Output:}} \textit{幸福\ 和平\ 惬意\ 放松\ 亲密\ 安分\ 随意}
~\\
\hrule
~\\

\end{CJK*}
\subsection{Result}
\begin{table}[H]
    \centering
    \renewcommand{\arraystretch}{1.5}
    \caption{The MRR scores of each model.}
    \resizebox{\linewidth}{!}{
    \begin{tabular}{cccc}
    \hline
    \multirow{2}*{Model} & \multicolumn{3}{c}{MRR\ Score} \\
    \cline{2-4}
    ~ & Total & Monosyllabic Words & Polysyllabic Words \\
    \hline
       ChatGPT-4o  & 41.90 & 45.61 & 40.09 \\
       Llama3-8b-Chinese  & 3.21 & 4.68 & 2.50  \\
   \hline
       Qwen2.5-14b  & 11.45 & 9.90 & 12.19 \\
       Qwen-FT & 19.16 & 20.80 & 19.37 \\
   \hline
        \textbf{CNMBERT}  & \textbf{61.53} & \textbf{74.30} & \textbf{55.35} \\
       $-$MoE & 59.70 & 72.29 & 53.61 \\
       $-$Multi-Mask & 34.49 & 49.97 & 27.01 \\
    \end{tabular}
       
    }
    \label{tab:my_label}
\end{table}
\begin{CJK*}{UTF8}{gbsn}

\begin{table*}
    \centering
    \renewcommand{\arraystretch}{1.6}
    \caption{Some results of Qwen-FT and CNMBERT.}
     \resizebox{\linewidth}{!}{
    \begin{tabular}{|c|c|c|c|c|c|}
    \hline
    \multirow{2}*{Examples} & 你快去给魔理沙看\textcolor{red}{\textbf{b}}吧. & \textcolor{red}{\textbf{bhys}}，我今天迟到了. &
    \textcolor{red}{\textbf{plq}}里吵起来了. & 落后了就要\textcolor{red}{\textbf{ad}}.  &  \textcolor{red}{\textbf{nsdd}}，但是...\\
    ~ & \textit{You can take Marisa to \textcolor{red}{\textbf{[b]}}.} & \textit{\textcolor{red}{\textbf{[bhys]}}, I'm late for work today.} &  \textit{There's a fight going on in the \textcolor{red}{\textbf{[plq]}}. } &  \textit{If you fall behind, you get \textcolor{red}{\textbf{[ad]}}.} 
    &  \textit{\textcolor{red}{\textbf{[nsdd]}}, but...}\\
    \hline
    \multirow{3}*{Qwen-FT} & 布(cloth) & 卑微少女(humble girl) & 棒球场(baseball field) & 复制(copied) & 不太确定(not sure) \\
    ~ & 表(clock) & \textcolor{blue}{\textbf{不好意思(sorry)}} & 排球场(volleyball field) & 挨饿(starve) & 非常确定(pretty sure)\\
    ~ & \textcolor{blue}{\textbf{病(illness)}} & 抱歉哈(I'm sorry) & 乒乓球(ping-pong ball) & \textcolor{blue}{\textbf{挨打(beaten)}} & 非常可能(extremely certain)\\
    \hline
     \multirow{3}*{CNMBERT} & \textcolor{blue}{\textbf{病(illness)}} & \textcolor{blue}{\textbf{不好意思(sorry)}} & \textcolor{blue}{\textbf{评论区(comment area)}} & 爱的(loved) & 你是懂得(you know)\\
     ~ & 包(bag) &  不好要是(not good if) & 评论群(comment group) &  \textcolor{blue}{\textbf{挨打(beaten)}} & 你说大道(you say road)\\
     ~ & 报(newspaper) & 毕后于是(after that) & 评论圈(comment circle) & 安迪(andy) & 你说到的(you said it)\\
     \hline
     \multirow{2}*{Target} & \multirow{2}*{\textcolor{blue}{\textbf{病(illness)}}} &\multirow{2}*{\textcolor{blue}{\textbf{不好意思(sorry)}}} & \multirow{2}*{\textcolor{blue}{\textbf{评论区(comment area)}}} & \multirow{2}*{\textcolor{blue}{挨打(beaten)}} & \multirow{2}*{\textcolor{blue}{你说得对(you are right)}}\\ 
     ~ & ~ & ~ & ~ &  ~ & ~\\
    \hline
    \end{tabular}
    }
    \label{tab:my_label}
\end{table*}
\end{CJK*}
\begin{table*}
    \centering
    \renewcommand{\arraystretch}{1.2}
    \caption{MRR@1, MRR@5, MRR@10 scores.}
     \resizebox{\linewidth}{!}{
    \begin{tabular}{c|ccc|ccc|ccc}
    \hline
    \multirow{2}*{Model} & \multicolumn{3}{c}{Total} & \multicolumn{3}{c}{Monosyllabic Words} & \multicolumn{3}{c}{Polysyllabic Words}\\ 
    \cline{2-10}
    ~ & MRR@1 & MRR@5 & MRR@10 & MRR@1 & MRR@5 & MRR@10 & MRR@1 & MRR@5 & MRR@10 \\
    \hline
    ChatGPT-4o & 37.05 & 41.52 & 41.81 & 39.83 & 45.08 & 45.47 & 35.69 & 39.78 & 40.03 \\
    Qwen-FT & 10.21 & 18.82 & 19.10 & 10.59 & 20.28 & 20.69 & 10.03 & 18.12 & 18.34 \\
    \hline
        \textbf{CNMBERT} & \textbf{51.86} & \textbf{60.39} & \textbf{61.22} & \textbf{64.75} & \textbf{73.59} & \textbf{74.30} & \textbf{45.63} & \textbf{54.01} & \textbf{54.90} \\
        $-$MoE & 49.74 & 58.56 & 59.41 & 62.05 & 71.56 & 72.29 & 43.79 & 52.28 & 53.18 \\
        $-$Multi-Mask & 26.70 & 33.37 & 34.21 & 39.18 & 48.85 & 49.97 & 20.67 & 25.90 & 26.60\\
    \hline

    \end{tabular}
    }
    \label{tab:my_label}
\end{table*}
In Table V, we present the overall MRR scores of different models on the test dataset, along with the MRR scores for predicting monosyllabic words and polysyllabic words separately. We found that the ability to predict pinyin abbreviations of GPT models without fine-tuning is quite limited. 
After fine-tuning, the performance of Qwen2.5-14b has significantly improved, but it still does not surpass CNMBERT. As autoregressive models with unidirectional attention, the GPT models typically perform poorly on fill-mask tasks that require obtaining information from both sides of the context. Additionally, GPT models do not utilize the MLM pretraining task, which further diminishes their performance.
% Additionally, they may exhibit significant hallucination, which sometimes leads to answers that do not conform to the constraints of pinyin abbreviations. 
% For Llama3-Chinese-8B-Instruct, even if it has been fine-tuned on Chinese datasets, the scarcity of Chinese resources in its original training datasets leads to poor adherence to prompts. As a result, it often generates a significant amount of irrelevant English content. 
For CNMBERT with multi-mask strategy, the experimental results indicate that it can effectively infer the meaning of pinyin abbreviations, with its performance significantly outperforming the others. Additionally, the results of the ablation experiment show that the multi-mask strategy and the inclusion of MoE layers significantly enhance its performance. 
However, when there is too little contextual information or the pinyin abbreviation sequences are too long, CNMBERT's performance may experience a sharp decline.
The examples of the models' outputs are shown in Table VI.

\par
We also evaluated our models' performance using MRR@1 (Accuracy), MRR@5, and MRR@10 scores, as shown in Table VII. From the results, we observed that some correct answers did not always appear in the first position but were instead found in the second or third positions.
% Moreover, because polysyllable consists of multiple Chinese characters. 
% As the length of the pinyin abbreviations increases, the number of possible outcomes grows exponentially, making it challenging for the model to make predictions. 
% Models like Qwen-FT, as autoregressive models, can generate text step by step, whereas autoencoder models like CNMBERT lack this capability, resulting in weaker performance when generating long texts. As shown clearly in Fig. 3, BERT's performance declines more sharply as the length of the predicted word increases.

Moreover, polysyllabic words consist of multiple Chinese characters, and as the length of the pinyin abbreviation increases, the number of possible outcomes grows exponentially. This significantly increases the difficulty of making accurate predictions. While autoregressive models like Qwen-FT can generate text sequentially, autoencoder models like CNMBERT lack this capability, leading to weaker performance on longer pinyin abbreviation sequences. As shown in Fig. 3, CNMBERT's performance declines more sharply as the length of the target word increases than GPT models.
\par

To address this issue, we tested whether the model could accurately predict the correct pinyin for both monosyllabic and polysyllabic words from its pinyin abbreviation. We continued to use MRR as the evaluation metric, and the results are shown in Fig. 4. We found that for our model, the MRR score of predicting the pinyin of words was higher than the score of predicting the words themselves. This indicates that the model can vaguely infer the pinyin of corresponding Chinese characters from the pinyin abbreviations. However, accurately predicting the correct words remains a challenging task for the model, highlighting a potential direction for further optimization.
\begin{figure}
    \centering
    \includegraphics[width=1.0\linewidth]{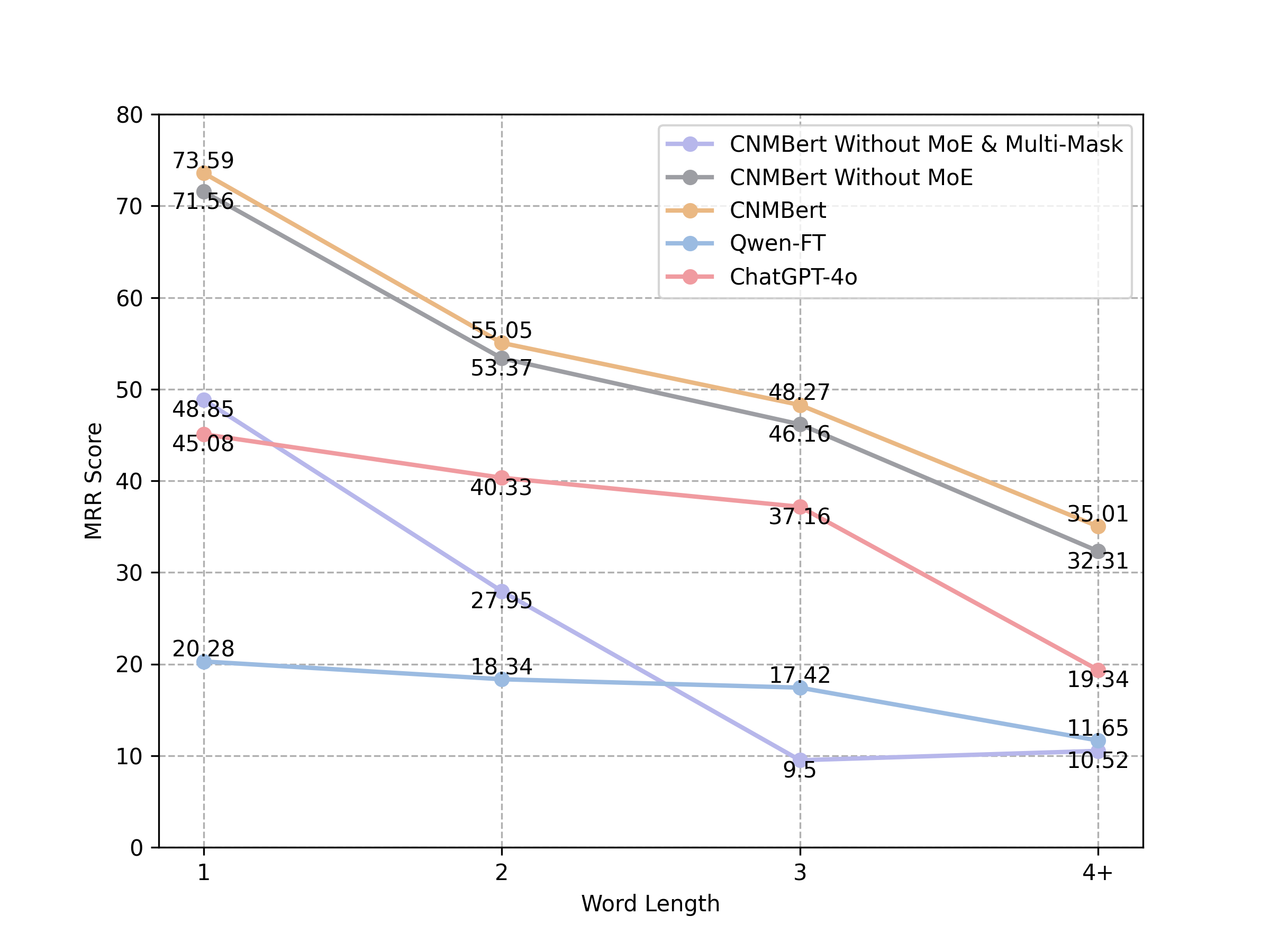}
    \caption{The scores using MRR@5 for predictions of words with different lengths.}
    \label{fig:enter-label}
\end{figure}
\begin{figure}
    \centering
    \includegraphics[width=1.0\linewidth]{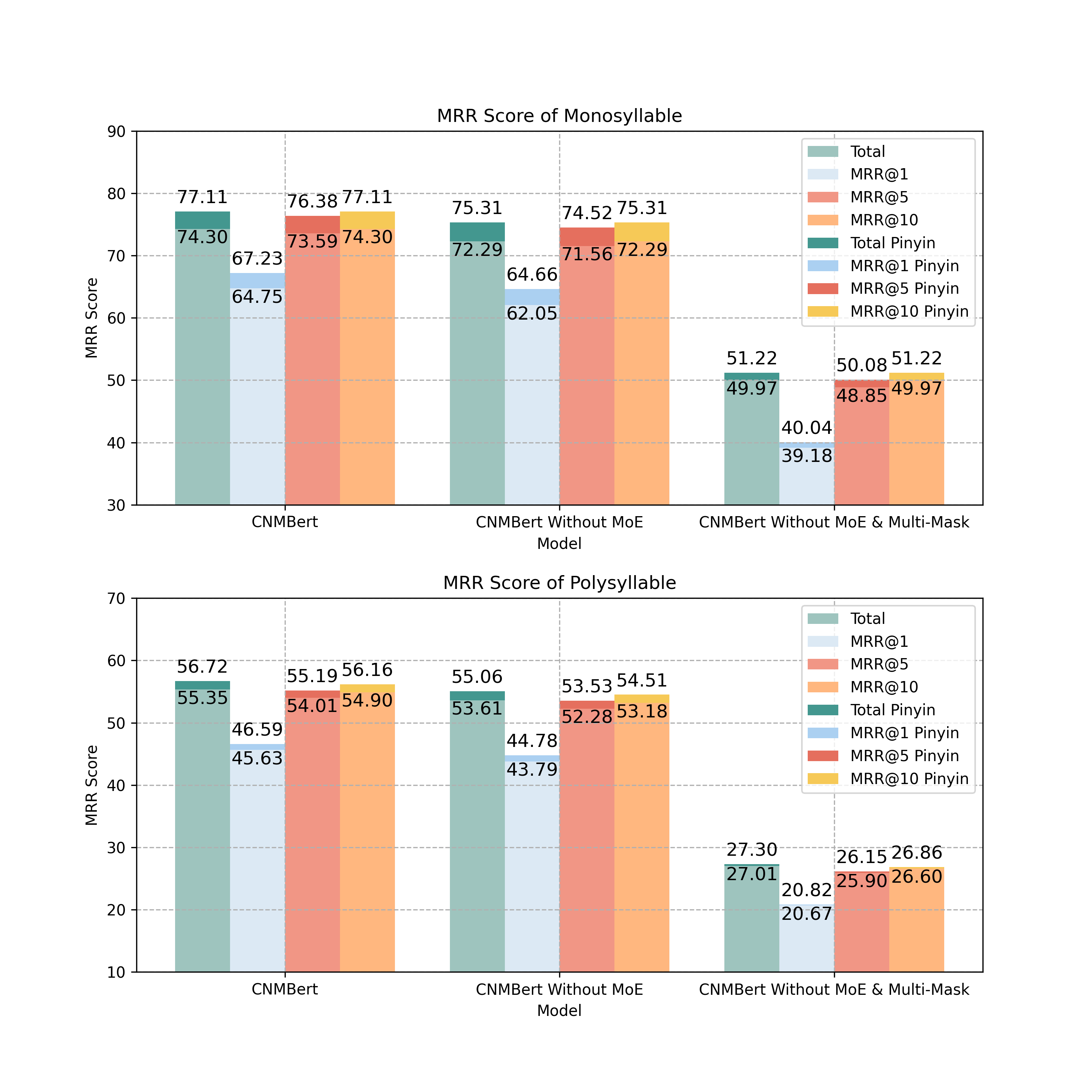}
    \caption{Results of predict monosyllabic and polysyllabic words’ pinyin.}
    \label{fig:enter-label}
\end{figure}

\par
In the final part of the experiment, we evaluated the queries per second (QPS) of different models. For this evaluation, we utilized the test dataset. Moreover, all models are loaded using the Transformers package \cite{b21} and run on a test environment with one RTX 3090 Ti GPU. As shown in Table VIII, in the test environment, CNMBERT without MoE outperforms in terms of memory usage, indicating its advantage in performance and resource-sensitive application scenarios. It also demonstrates potential for integration into tools used for CSC.

\begin{table}
    \centering
    \renewcommand{\arraystretch}{1.2}
    \caption{The QPS and memory usage of each model. Qwen-FT uses shared memory.}
    \resizebox{\linewidth}{!}{
    \begin{tabular}{|c|c|c|c|}
    \hline
       Model & QPS & Memory Usage (FP16) & Model Size \\
    \hline
        Qwen-FT & 0.075 & 28.9GB & 14.8B \\
    \hline
        CNMBERT & 3.20 & 0.8GB & 329M \\
        $-$MoE & 12.56 & 0.4GB & 131M \\
        $-$Multi-Mask & 12.77 & 0.4GB & 131M \\

    \hline
    \end{tabular}
    }
    \label{tab:my_label}
\end{table}
\begin{figure*}
    \centering
    \includegraphics[width=1\linewidth]{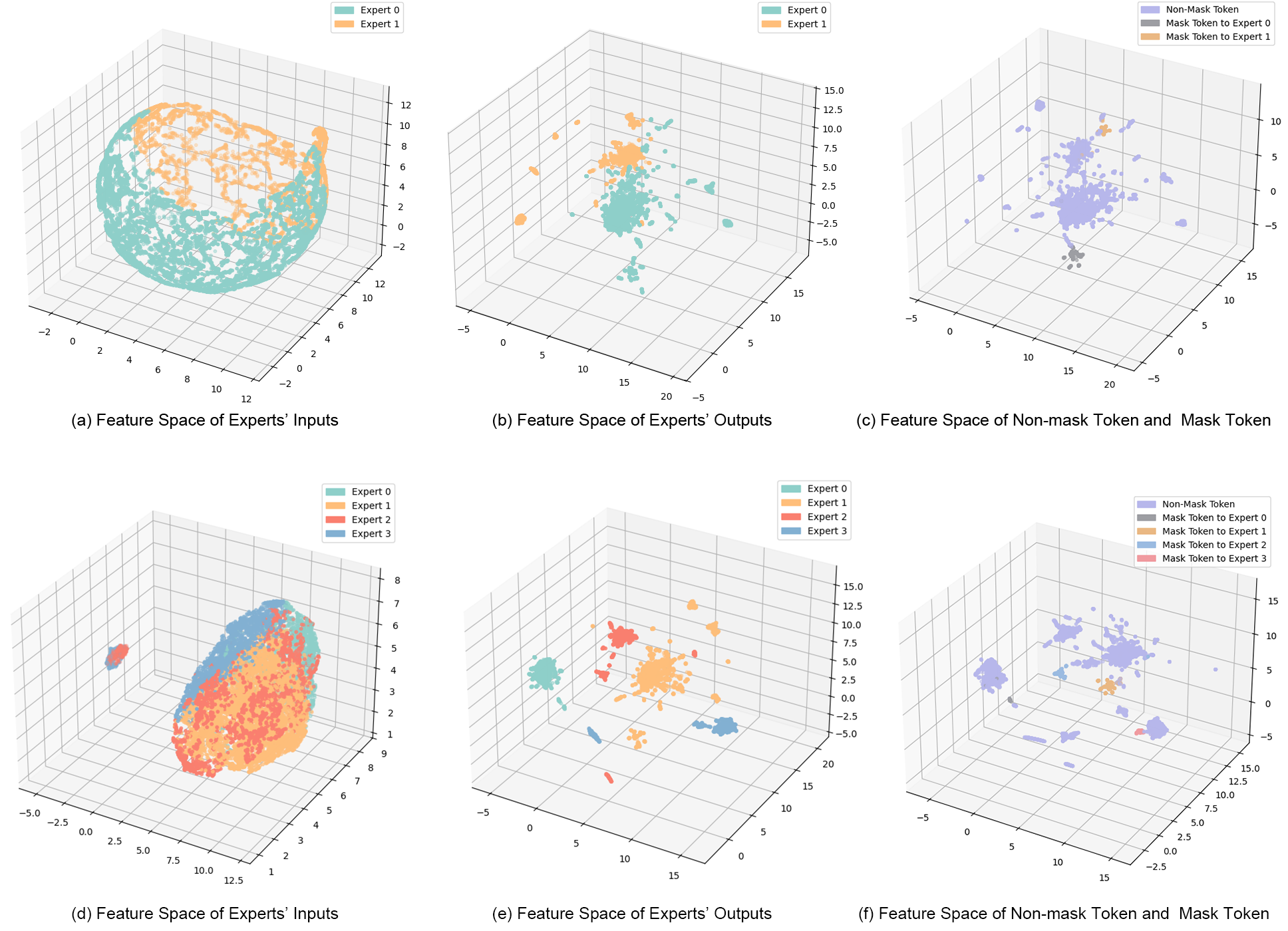}
    \caption{The feature space of experts in layer 3(a-c) and layer 7(d-f).}
    \label{fig:enter-label}
\end{figure*}
\subsection{Analysis of MoE layers}
To validate the effectiveness of the MoE layers, we applied UMAP \cite{b26} to reduce the dimensionality of the input and output features for each expert. This enabled us to determine whether different experts in the MoE layers had learned to process data with distinct feature characteristics. Thus, we analyzed the input and output features of each expert in the MoE layers. For MoE layers closer to the input, we additionally examined the feature differences between mask tokens and other tokens, as well as the experts to which mask tokens were routed.

In Fig. 5, we observed a clear division of labor among the experts when processing tokens. Specifically, certain experts demonstrated a tendency to handle input tokens with specific feature patterns, while others focused on distinct feature subsets. Furthermore, the tokens processed by individual experts exhibited clustering tendencies, indicating that the experts possess particular preferences or functional specializations within the output feature space.
\par
The mask tokens exhibit a distinct clustering trend within the feature space, with their distribution markedly different from that of regular tokens, especially in the layers closer to the input. This suggests that the model handles mask tokens with differentiated feature representations. In addition, different experts process mask tokens belonging to separate clusters, indicating that the experts are capable of capturing the feature variations among mask tokens and performing targeted processing accordingly.

\par
\section{Conclusion and Future Work}
In this paper, to address the problem of converting pinyin abbreviations to Chinese characters, we improved the traditional BERT model architecture and masking logic, proposing CNMBERT. Using our constructed test dataset, we validated that this model outperforms other GPT models and the baseline model in terms of accuracy for the pinyin abbreviation to character conversion task. CNMBERT introduces a novel multi-mask strategy and MoE-based approach, which can be adapted for solving the pinyin abbreviation problem and can help people better understand the meaning of pinyin abbreviations, offering potentially improved solutions for tasks such as named entity recognition and sentiment analysis. 
\par
% Future work will focus on improving accuracy by training the model on larger datasets. Additionally, we will optimize the model's structure to enhance performance further as well as apply the model to other languages, such as English.
Although our model achieves better performance in converting pinyin abbreviations to Chinese characters compared to other models, its performance still experiences a sharp decline when faced with a lack of contextual information or long pinyin abbreviations.
To improve the performance of this model, future work will be performed to explore enhanced strategies for resolving pinyin ambiguity through using the multi-mask strategy and expanding the dataset. The proposed CNMBERT architecture will be also applied to the task of translating abbreviations in other languages, such as translating English acronyms.

\vspace{12pt}

\end{document}